# MudrockNet: Semantic Segmentation of Mudrock SEM Images through Deep Learning

Abhishek Bihani[1], Hugh Daigle[1], Javier E. Santos[1], Christopher Landry[2], Maša Prodanović[1], Kitty Milliken[3]

[1]Hildebrand Department of Petroleum and Geosystems Engineering, The University of Texas at Austin, Austin, Texas, USA; [2]Center for Subsurface Energy and the Environment, University of Texas at Austin, Austin, Texas, USA; [3]Bureau of Economic Geology, University of Texas at Austin, Austin, Texas, USA[1]

## Abstract

Segmentation and analysis of individual pores and grains of mudrocks from scanning electron microscope images is non-trivial because of noise, imaging artifacts, variation in pixel grayscale values across images, and overlaps in grayscale values among different physical features such as silt grains, clay grains and pores in an image, which make their identification difficult. Moreover, because grains and pores often have overlapping grayscale values, direct application of threshold-based segmentation techniques is not sufficient. Recent advances in the field of computer vision have made it easier and faster to segment images and identify multiple occurrences of such features in an image, provided that ground-truth data for training the algorithm is available. Here we propose a deep learning SEM image segmentation model, MudrockNet based on Google's DeepLab-v3+ architecture implemented with the TensorFlow library. The ground-truth data was obtained from an image-processing workflow applied to scanning electron microscope images of uncemented muds from the Kumano Basin offshore Japan at depths < 1.1 km. The trained deep learning model obtained a pixel-accuracy about 90%, and predictions for the test data obtained a mean intersection over union (IoU) of 0.6591 for silt grains and 0.6642 for pores. We also compared our model with the random forest classifier using trainable Weka segmentation in ImageJ, and it was observed that MudrockNet gave better predictions for both silt grains and pores. The size, concentration, and spatial arrangement of the silt and clay grains can affect the petrophysical properties of a mudrock, and an automated method to accurately identify the different grains and pores in mudrocks can help improve reservoir and seal characterization for petroleum exploration and anthropogenic waste sequestration.

[1] AB carried out the conventional segmentation for ground truth image labeling, trained the model and drafted the manuscript. HD supervised the work, acquired funding, and helped draft the manuscript. JES helped in model training and evaluation and helped draft the manuscript. CL created the conventional segmentation code used for ground truth image labeling. MP helped in implementing the conventional segmentation code and helped draft the manuscript. KM acquired the mudrock samples and prepared the SEM images used for the project.  All authors have read, reviewed, and approved the manuscript.

## 1.0 Introduction

The dominant sedimentary rocks on Earth, mudrocks are composed of silt- and clay-size particles, that typically include clay minerals, quartz, feldspar, and carbonate, that are smaller than 63 μm (Macquaker and Adams, 2003; Lazar et al., 2015b). These lithologies are important as capillary seals over hydrocarbon accumulations (Schowalter, 1979; Schlömer and Kroos, 1997), as caprocks for carbon capture and storage (Li et al., 2005), and as unconventional oil and gas reservoirs (e.g., Bustin et al., 2008). At higher concentration, silt grains provide the support framework for the mudrocks, and at lower concentration they shelter the larger pores from compaction (Oertel, 1983; Yang and Aplin, 2007; Desbois et al., 2009; Schneider et al., 2011; Pommer and Milliken, 2015). The size, concentration, and spatial arrangement of silt- and clay-size grains affect the petrophysical properties of in mudrocks, thereby affecting fluid transport behavior (Potter et al., 2005; Lazar et al., 2015a; Bihani and Daigle, 2019). Hence, scanning electron microscope (SEM) images of mudrock samples can serve an effective tool for improved reservoir and caprock characterization.

A typical workflow for filtering and segmenting microstructural images of rocks consists of steps to identify the individual features, each requiring the input of the user, for example fixing the segmentation threshold pixel values for each image. However, with recent advances in machine learning, it is possible to reduce the number of steps and the degree of user intervention required for feature detection or image segmentation. Other researchers have recently used machine learning techniques successfully for feature detection in high-resolution SEM images of rocks. Tian and Daigle (2019) used automated object detection algorithms on high-resolution backscattered electron images and energy-dispersive X-ray spectroscopy (EDX) images to characterize the location and size of microfractures and their preferential association with particular minerals in shales. However, the object detection algorithm is only suitable for recognizing features which have specific shapes and are limited to only a few occurrences in an image. If the algorithm is applied to an SEM image containing hundreds of pores and grains, the bounding boxes cannot capture all the features accurately. Maitre et al. (2019) used a simple linear iterative cluster segmentation for automating the recognition of minerals like plagioclase, ilmenite, monazite, magnetite from high-resolution red-green-blue (RGB) SEM images of sand grains. They compared different machine learning classifiers like random forest, k-nearest neighbors, and decision trees on the RGB color channels and found that random forest had the highest global accuracy (0.89). The method exploited the differences in mineral colors in similarly sized grains for grain recognition but cannot be applied for segmenting grayscale SEM images. Tang and Spikes (2017) used high-resolution EDX images of shale samples as inputs in a neural network to detect calcite, feldspar, quartz, kerogen, and clay/pore, but were unable to distinguish between clays and pores. Wu et al. (2019) used feature extraction followed by the gradient boosting and random forest classifiers for segmentation of high-resolution grayscale shale SEM images. They found that the random forest classifier gave the best results for detecting the four types of features; pores/fractures, rock matrix (quartz, clay, calcite), pyrites and kerogen. However, they did not identify the different types of grains. Thus, while color-coded data in SEM images makes it relatively easy to detect different minerals, identification of individual pores and grains of specific types and sizes from SEM images, such as those in a mudrock, is a more challenging task.

## 2.0 Background

In the field of computer vision, the task of assigning labels pixel by pixel in an image is termed semantic segmentation (Chen et al., 2018). Semantic segmentation requires identifying irregular object outlines in an image, and therefore, has stricter accuracy requirements than conducting simple image-level classification or bounding-box based object detection in an image (Liu et al., 2019).We propose using it to detect pores as well as particular types of grains (e.g., silt) in the images, with sufficient amount of data from raw images and ground truth are provided for training. Moreover, whereas semantic segmentation can identify different object classes (pores and silt grains), it is possible to enhance the segmentation to instance-level segmentation, which entails detecting each instance of grain and pore in the image, and assigning a unique identifier for each instance. This can also be done in a post-processing step, wherein all the unconnected grains and pores that have been detected are assigned a unique identifier to allow further analysis.

Convolutional neural networks (CNN) use shared-weight architecture and are effective for image segmentation, since they have different types of layers whose parameters can be trained by data for performing a particular task (Liu et al., 2019). Long et al. (2015) introduced fully convolutional neural networks to image segmentation by replacing fully connected layers (all neurons in a layer connected to all neurons in previous layer) with fully convolutional layers (neurons in a layer are connected to a small region of the previous layer). As shown in Figure 1A, an input image passed through multiple convolutional layers of the encoder (black arrows), undergoes successive reductions in the height and width of the convolved image representations (feature maps). Thus, the encoder converts the input data into a feature map by multiplying (convolving) the original image with a smaller layer (kernel/filter) to identify important features in the image which would not be easily identifiable to humans (Goodfellow et al., 2016). The first output feature map is then used as the input in the next convolution with another kernel and the process continues until reaching the bottleneck layer (blue dotted line), which has the highest compressed representation of the input data. Thereafter, a deconvolution network (decoder) can be used to up-sample the image through interpolation across successive layers to recover the feature map back to the original image size (Noh et al., 2015).

Ronneberger et al. (2015) built on this and modified it to create the U-net architecture, which worked with fewer training images but improved the results. It uses a symmetric encoder-decoder architecture with a large number of feature channels in the encoder part, which helps in preserving the information passed to the high-resolution layers, and the decoder layers are connected to the corresponding encoder layers to improve the resolution of predicted image. Since that time, further advances in semantic segmentation have occurred, including DeepLab, which is a state-of-the-art supervised segmentation model created and released as open source by Google (Chen et al., 2014). It was first released as DeepLab-v1 (Chen et al., 2014) and was followed by multiple improvements; DeepLab-v2 (Chen et al. 2017a), DeepLab-v3, (Chen et al. 2017b), DeepLab-v3+ (Chen et al. 2018).

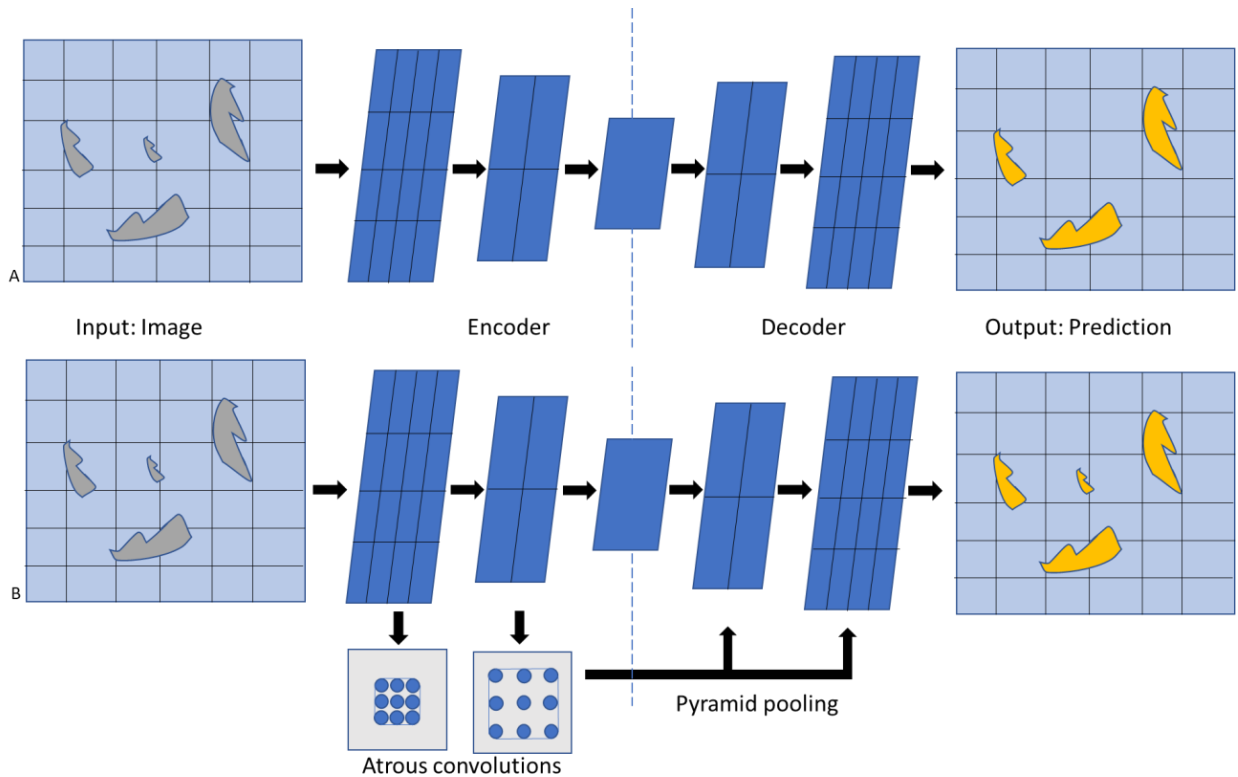

Figure 1 - Comparison of A) Normal encoder-decoder architectures, B) DeepLab-v3+ architecture with atrous convolutions and pyramid pooling. Modified after Chen et al., 2018. Convolutions between successive feature maps are shown by black arrows, and grid lines indicate the number of pixels in the particular image feature map.

MudrockNet is based on DeepLab-v3+, which also uses an encoding-decoding architecture to extract information using convolutional layers. However, whereas applying an encoder-decoder architecture can decrease computational times for neural network training and capture sharp object boundaries, the repeated down-sampling may miss rich semantic information when the resolution of training layers is smaller than the features to be identified (Chen et al., 2018). The architecture can also cause problems while recognizing objects at multiple scales. For example, if an SEM image of a mudrock has pores of multiple sizes, the smallest pores consisting of just few pixels may not be identified by models using conventional encoding-decoding architecture (Figure 1A). The DeepLab-v3+ architecture overcomes these issues by combining two techniques, atrous convolutions and spatial pyramid pooling (atrous spatial pyramid pooling) (Figure 1B). Applying parallel atrous (dilated) convolutions effectively increases the field of view, incorporating multi-scale context, because the atrous filters prevent the down-sampling operations in the last few layers by inserting holes between the filter weights (Chen et al., 2018). The spatial pyramid pooling uses parallel versions of the same image at different scales (feature maps) which are later combined, and when used along with atrous convolutions and encoder-decoder modules can provide better results for semantic segmentation.

Therefore, we trained the MudrockNet model to identify pores as well as silt-size grains from the SEM images. This model can be used for segmentation of other mudrock SEM images, and the individual pores and grains can be used for further analysis.

## 3.0 Materials and Methods

Figure 2 describes the workflow for semantic segmentation using MudrockNet and result comparison. The process consists of three parts: data pre-processing, model training and testing, and data post-processing.

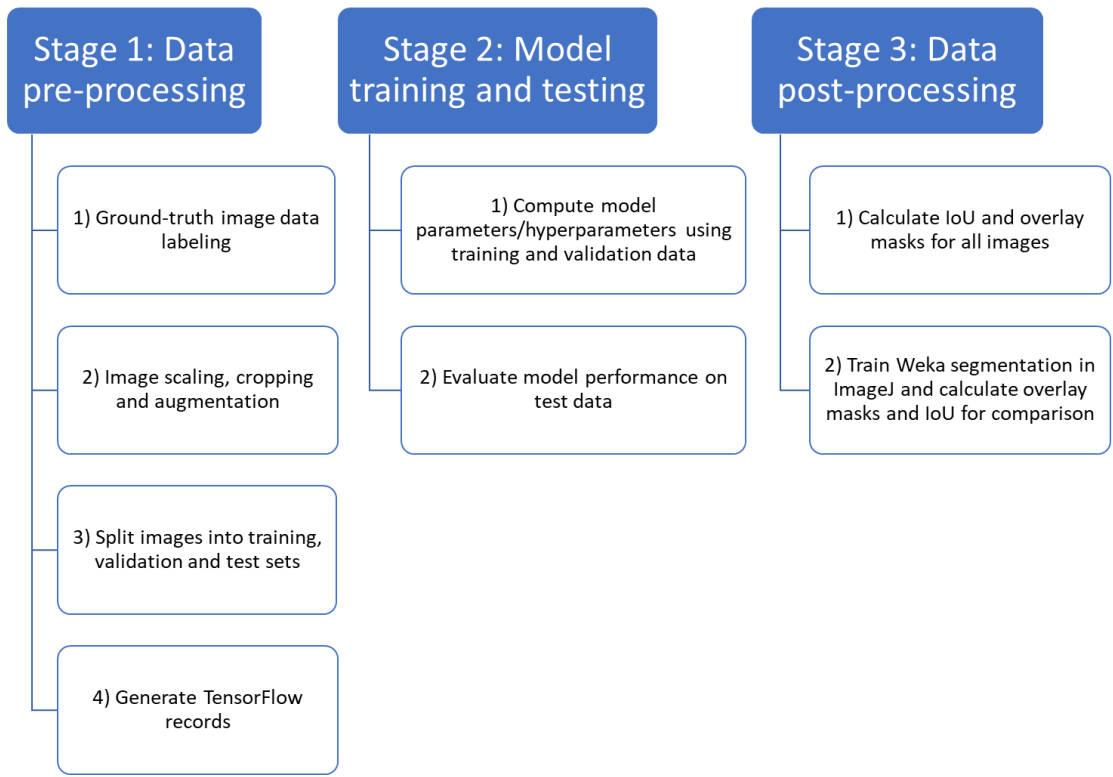

Figure 2 - Workflow for semantic segmentation and comparison of results

The SEM images for the model were obtained from uncemented mud samples in the Kumano Basin, which is a large forearc basin in the Nankai trough, offshore Japan (Moore et al., 2009). These core samples were acquired at Site C0002 during Integrated Ocean Drilling Program (IODP) expeditions 315 and 338 at depths < 1.1 km below the sea floor (Milliken et al., 2016). Forty-nine SEM images from five core samples at different depths were used for the study. The sample surfaces for pore imaging were prepared by Ar-ion cross-section polishing and were coated in 6 nm iridium in a Leica EM ACE600 to minimize surface charging effects. The images were produced by secondary electron detection using the FEI Nova-NanoSEM 430 scanning electron microscope by a mixed signal of backscattered electron and secondary electron detection to reduce charging effects (Nole et al., 2016). All the images were scanned at a machine magnification of either 15,000 X or 40,000 X resulting in horizontal field widths (HFW), i.e. image widths of 20 µm or 7.5 µm respectively.

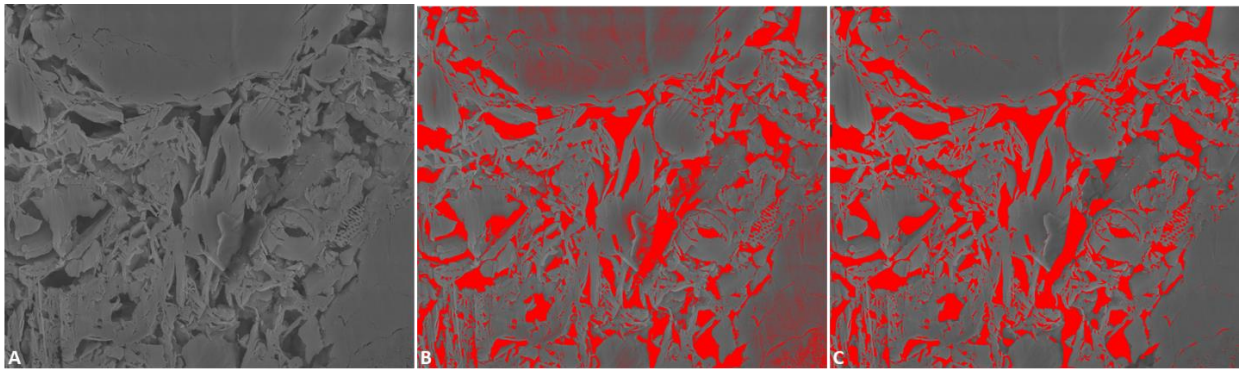

Figure 3 - SEM images: A) Original; B) Segmentation without filtering; C) Segmentation after filtering (predicted pores are shown in red).

An image processing workflow was applied to SEM images following Landry et al. (2017) to obtain the ground truth data for the model. This is necessary since simple threshold-based segmentation techniques without filtering cannot properly separate grains and pores as they can have overlapping grayscale values. For example, on applying simple segmentation to the image in Figure 3A results in wrongly predicting darker grain pixels as pores (Figure 3B). Therefore, the first step in the workflow was to apply a median filter to smooth the image to the targeted spatial scale of features to be segmented, followed by a combination of top-hat and bottom-hat filter to increase the local contrast. After smoothing and contrast enhancement simple, a grayscale threshold value is applied to separate the image into pores and grains without overlaps (Figure 3C). The image segmentation was done at different spatial scales, to capture pores of all sizes and the final image was merged together, similar to pyramid pooling. The silt-size grains (equivalent circular diameter > 2 µm) were identified by eroding and dilating the grain component multiple times as specified by the user. This dataset of all original and segmented images (silt, clay, pores) used in this study using this conventional segmentation algorithm is available at Bihani et al. (2020).

To improve model training, all images are scaled to the same magnification of 15,000X. Thereafter, to maintain consistent image size and increase the total number of training images, we divided the SEM images into multiple equal parts, each equal to the height and width of the smallest scaled SEM image (400 x 343 pixels) as shown for an example image in Figure 4A, B, C. Horizontal (Figure 4D) and vertical (Figure 4E) flipping of the images was carried out for data augmentation by following the methods of Tian and Daigle (2018), resulting in a total of 1785 images. The images (original and corresponding ground truth) were then split into training (~70%), validation (~15%), and test (~15%) datasets.

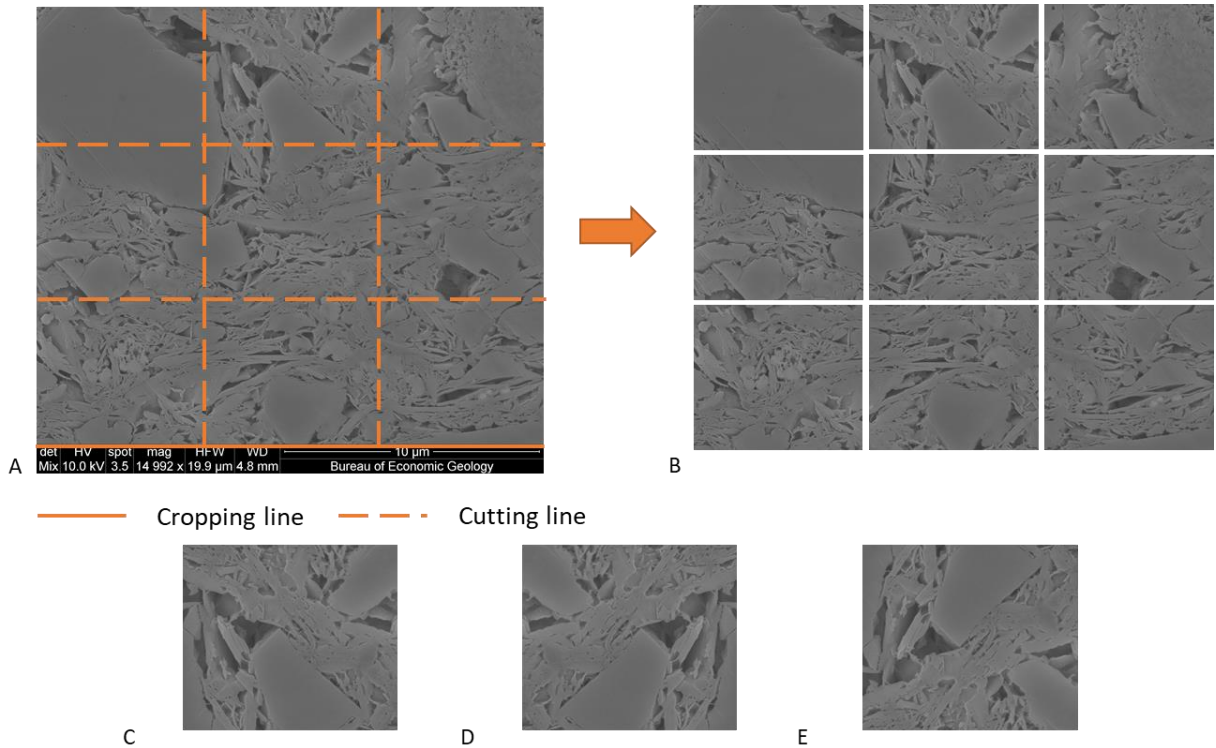

Figure 4 - A) Original image (magnification 15,000 X), B) After cutting and cropping, C) Sample input image, D) Horizontally flipped image, E) Vertically flipped image

The MudrockNet model was trained using a NVIDIA GeForce GTX 1070 GPU with 8 GB memory for the three classes: silt, pore, and clay. The training was aided by transfer learning from a pre-trained ResNet-101 model and was continued for 50 epochs until the loss became constant, signaling occurrence of a local/global minima. We used TensorFlow's sparse softmax cross entropy loss with addition of weight decay, and the momentum optimizer for loss gradient calculations. The particular loss function is suitable for mutually exclusive classification of multiple classes (Wang et al., 2019), and the inclusion of weight decay regularization helps prevent overfitting by penalizing large weights (Schmidhuber, 2015). The momentum optimizer helps the stochastic gradient descent navigation during training by adding a fraction of the direction from the previous steps which reduces irrelevant oscillations (Sutskever et al., 2013). The training was evaluated by two metrics: pixel accuracy and intersection over union (IoU) for both pores and silt grains. IoU or the Jaccard index (Figure 5) is defined as the area of intersection (common pixels) between the prediction and the ground truth divided by the area of their union (pixels present in both images) (Tian and Daigle, 2018). The pixel-accuracy is defined as the number of pixels labeled correctly over the total number of pixels in the image. The IoU is preferred over pixel accuracy since it is not affected by class imbalance (Rahman et al., 2016). Thereafter, the trained MudrockNet was tested on the test dataset for an unbiased evaluation of its ability to predict the features in different cases. We found that using transfer learning gave better mean IoU results on the test data (silt grains- 0.6591, pores- 0.6642), when compared with the results for a model trained without pre-trained weights (silt grains- 0.4671, pores- 0.6592). The predicted segmentations are compared with ground-truth data by calculating the mean IoU for silt grains and pores for each image, and typically using a low threshold, IoU values > 0.5 can be considered true positive (Li et al., 2017). The overlay masks of the MudrockNet predictions and the ground truth (conventional segmentation algorithm) predictions over the raw image are also calculated for visual inspection.

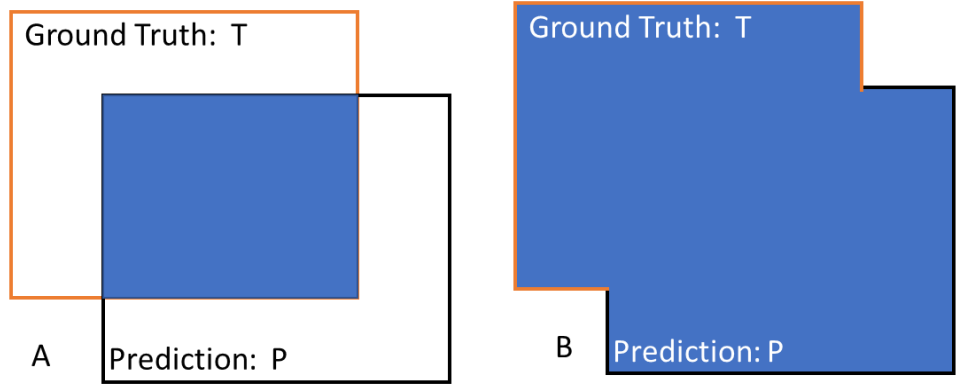

Figure 5 - A) Intersection area of ground truth and prediction, B) Union area of ground truth and prediction. Modified from Tian and Daigle (2018).

The trainable Weka segmentation with a random forest classifier in ImageJ (Arganda-Carreras et al., 2017) was also used to perform image segmentation, to compare the trained model's performance with other machine-learning methods. Random forest is an ensemble tree-based learning algorithm that averages multiple decision trees from randomly selected subsets of training data to build class predictions (Liaw & Wiener, 2002). The algorithm was selected since it has been successfully applied for problems in rock SEM images (Maitre et al., 2019; Wu et al., 2019), and being an ensemble method helps increase the accuracy while controlling overfitting. The Weka model was trained on 30 images from the training set with multiple samples for each class (silt, clay, pore). The training used edge-detection filters like the Sobel and Hessian filter along with a FastRandomForest classifier, which is a multi-threaded version of the random forest classifier using 200 trees and 2 random features per node. The trained Weka classifier was then applied on the test data to compare predictions using overlay masks and IoU values.

## 4.0 Results and Discussion

The error rate of the model on the training set indicates the model bias and the difference in the performance between the training and the test or validation set (generalization error) indicates the model variance. A high bias is likely to occur when the model algorithm size or complexity is insufficient for the problem and can cause underfitting, and conversely, a high variance is likely when a complex model has learned the training data too well to be applied widely by overfitting. While both contribute to the total error, the bias can decrease with increasing model complexity, while the variance can decrease on increasing the size of the dataset (Ng, 2019). Figure 6A shows the training and testing metrics of the MudrockNet model. The network training was stopped after 50 epochs, once the training and validation loss became constant (values 13.25 and 13.52 respectively), and the training and validation pixel-accuracy reached a plateau (values 0.9205 and 0.8898 respectively). It is seen that overfitting has not occurred since the training loss in the learning curve (Figure 6A) remains smaller than the validation loss. The mean IoU values for the individual classes (grains and pores) for the training, validation and test datasets are shown in Table 1, and at an image level are seen in Figures 6B and C.

| Mean IoU values | Training set | Validation set | Test set |
|---|---|---|---|
| **Silt grains** | 0.71696 | 0.64431 | 0.65917 |
| **Pores** | 0.68446 | 0.67441 | 0.66428 |

Table 1 - IoU values according to class

In Table 1, we can see from the comparison that the mean IoU values on the test and validation set are only slightly lower than the training set (low variance) and, thus, the model is able to generalize the mud segmentation problem. This indicates that the model can suitably identify the silt grains and pores in mudrock SEM image sets that are of similar quality and size to those used here. While a majority of the grain and pore IoU values in Figure 6B and C are larger than 0.5 (low threshold for true positive), the grain IoUs (Figure 6B) have a larger amount of scatter than the pore IoUs, especially in the validation and test data (Figure 6C). This may mean that MudrockNet is better trained to predict pores than the silt grains in new SEM images and is possible due to various factors like the variation in silt grain size, greater number of pores in the data, and due to the overlapping pixel values of silt grains and clay grains, which can lead to errors, especially for smaller silt grains.

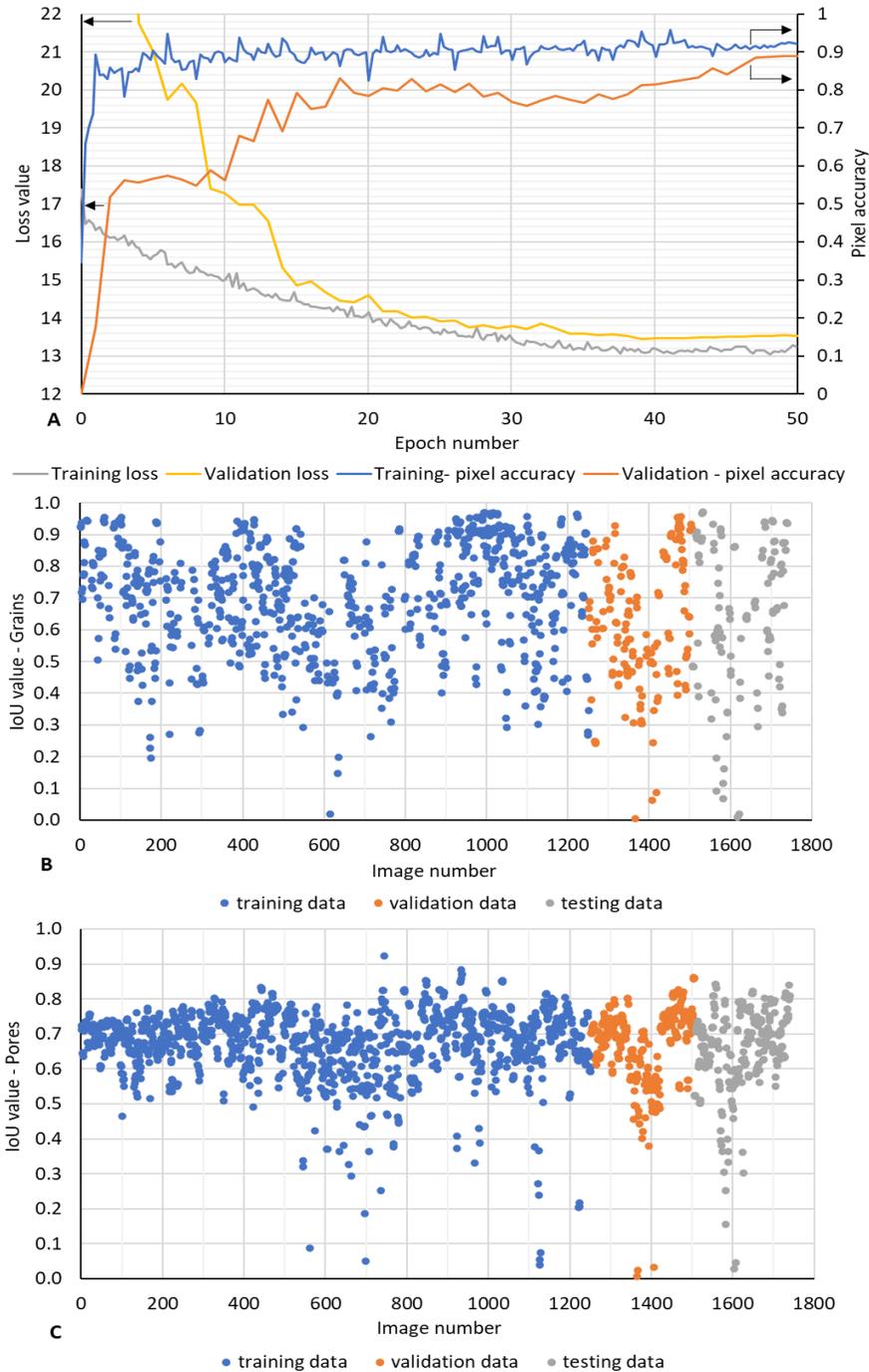

Figure 6 - Training and validation metrics for MudrockNet model: A) Training metrics (black arrows point to corresponding axes), B) Silt grain IoU values for individual images, C) Pore IoU values for individual images

Figure 7 shows the overlay mask of ground truth (conventional segmentation algorithm data) (A), MudrockNet model predictions (B), and trainable Weka model predictions (C), on four selected SEM images from the test set. The silt grains are in red, pores in green, clay in transparent color, and the truth images show a scale bar for reference. Other comparisons are available online at https://github.com/abhishekdbihani/MudrockNet/tree/master/dataset/data/masks_test.

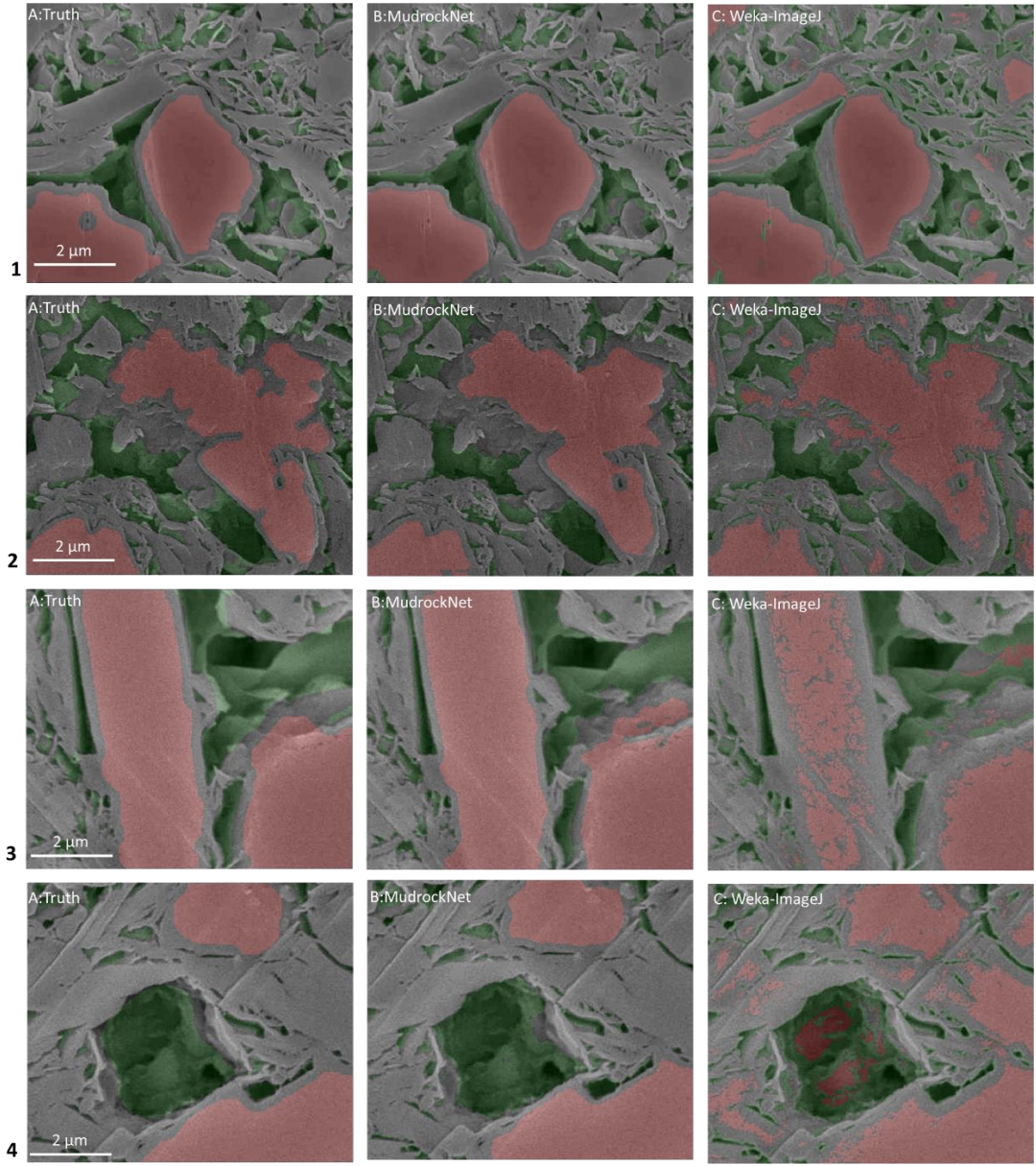

Figure 7 - Example SEM images of silt grains (red), pores (green), clay (transparent)- A) ground truth (conventional segmentation algorithm) data, B) MudrockNet model predictions, C) trainable Weka model predictions.

| IoU values | MudrockNet model | | Weka model | |
|---|---|---|---|---|
| Image | Silt grains | Pores | Silt grains | Pores |
| 6-1 | 0.892 | 0.729 | 0.702 | 0.581 |
| 6-2 | 0.822 | 0.655 | 0.665 | 0.667 |
| 6-3 | 0.889 | 0.667 | 0.578 | 0.707 |
| 6-4 | 0.881 | 0.816 | 0.543 | 0.497 |

Table 2 - Silt and pore IoU value comparisons from MudrockNet and Weka model for images in Figure 7

Table 2 shows the comparison of the IoU values for silt grains and pores by both the methods (MudrockNet and Weka model) for images shown in Figure 7. In Figure 7-1, we can see that the MudrockNet model predictions (1B) for both the silt size grains (red) and the pores of different sizes (green) match the ground truth image from conventional segmentation (1A) well, as confirmed from the IoU values. While the Weka model (1C) is also able to detect the silt grains, it is not able to adequate differentiate between silt and clay grains, as seen in the smaller red patches in the clay. In Figure 7-2, MudrockNet (2A) is able predict the entire silt grain (red in center) better than the ground truth image (2B) where the grain appears connected through smaller segments, thereby reducing the IoU. The Weka model (2C) is able to detect the larger structures, but overpredicts the silt grains and pores, possibly because of the darker image texture. Similar behavior is seen in Figure 7-3, where the predictions from MudrockNet (3A) are very similar to the ground truth (3B), with high IoU values, while the Weka model's (3C) silt predictions underpredict the actual silt grain pixels. In Figure 7-4, MudrockNet (4B) predicts silt grains and pores as seen in the ground truth image (4A) leading to a high IoU. The Weka model (4C) overpredicts silt pixels (red), in the clay grains and in the large central pore.

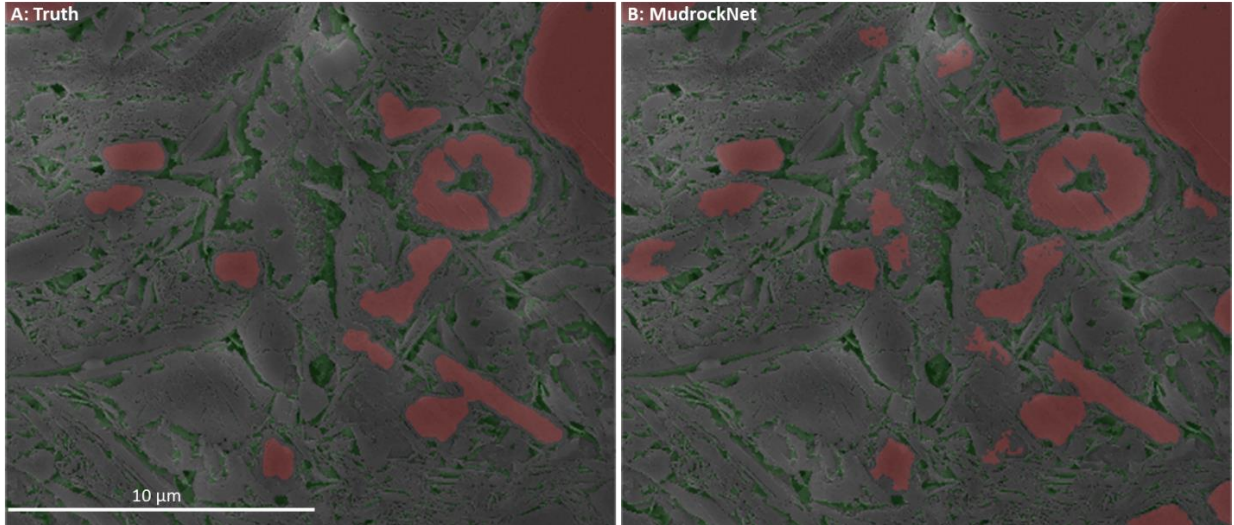

Figure 8 - Comparison of predictions for a large (magnification 15,000 X) test SEM image with silt grains (red), pores (green), and clay (transparent)- A) ground truth (conventional segmentation algorithm), B) MudrockNet model predictions.

Figure 8 shows the comparison of ground truth (A), and MudrockNet model predictions (B) for a large test SEM image (magnification 15,000 X) of 2048 x 1767 pixels. It can be seen that despite the image being larger than the images used for training, MudrockNet can identify the pores (green) and silt grains (red) with a grain IoU of 0.739 and pore IoU of 0.682. While there are some differences since MudrockNet may identify some smaller grains as silt size due to the larger size of the image, Figure 8 shows the capability of the model to segment SEM images larger than it was trained on. Additionally, while the prediction time for the conventional segmentation algorithm (with default values) took 7.767 seconds, the MudrockNet model's prediction took only 0.218 seconds, which is faster by an order of magnitude.

Thus, from the overall trends observed in the example images in Figure 7 and 8, and the corresponding IoU values in Table 2, we can infer that while the conventional filtering and segmentation algorithm used for ground truth images was able to successfully isolate the individual pores and silt grains well, the deep learning approach (MudrockNet) was able to obtain similar results with less user intervention, fewer steps, and faster. In contrast, the Weka model underpredicted or overpredicted the silt grains and pores, possibly due to limited training data and model limitations.

While in many cases the MudrockNet model results are close to those obtained from expert user-guided segmentation, the limited data availability despite data augmentation, reduces the ability of the model to correctly identify each and every instance of pore and silt grain in all mudrock images. Since the images have varying pixel values for the same feature in different images due to the SEM image generation process, the prediction models need to be able to be trained to capture this variation to prevent mislabeling. Moreover, the mean IoU observed for the silt grains was comparatively lower because of fewer training data compared to pores, and as the variable size silt grains in many cases have same pixel values as the clay grains. This can cause a difference in the predictions from the different methods, and they may have to be post-processed to remove grains below a specified size threshold. Finally, the uncertainty of the ground truth data due to the complex filtering and segmentation process can also result in predictions of pores and silt grains by the MudrockNet model which may have been overlooked in the conventional method.

## 5.0 Conclusion and future work

We propose an alternative method for filtering and segmentation using deep learning to identify pore and grain features from mudrock SEM images. The MudrockNet model is trained based on Google's DeepLab-v3+ architecture from the TensorFlow library, and the predictions for the test data obtain a mean IoU of 0.6591 for silt grains and 0.6642 for pores. Comparisons with the trainable Weka segmentation model in ImageJ showed that MudrockNet gave superior predictions. Once trained, deep learning algorithms can provide a segmentation result way faster than a conventional segmentation algorithm (such as the one here used for ground truth). This is specifically important as the imaging resolution improves, and large area SEM images become easily available.

The ability to detect pores and silt grains without user input and multiple stages can make the segmentation process more streamlined and easier to use. An automated method to accurately identify silt grains, clay-size particles, and pores can improve the characterization of mudrocks and may even lead to a better understanding of the role of mudrocks as reservoirs or seals for petroleum exploration or sequestration of radioactive waste or carbon dioxide. If more SEM image datasets are available along with the labeled ground truth data, then the model accuracy and the ability to better detect pores and grains can be further improved.

## 6.0 Data and code availability

The original mudrock SEM images used in this work are posted on Digital Rocks Portal, please see Milliken et al. (2016). The original images and the associated ground truth data is also available on Digital Rocks Portal (Bihani et al., 2020). Our trained MudrockNet model and the trainable Weka model for ImageJ are available in a GitHub repository at https://github.com/abhishekdbihani/MudrockNet.

## 7.0 Acknowledgments

The authors would like to thank the University of Texas at Austin for the support, and Equinor for funding this research through the Statoil (Equinor) Fellows program. Samples and data were provided by the Integrated Ocean Drilling Program (IODP). Funding for sample preparation and SEM imaging was provided by a post-expedition award (Milliken, P.I.) from the Consortium for Ocean Leadership. MudrockNet was implemented using the TensorFlow library in Python, and the implemented model and the architecture (DeepLab-v3+) has been modified from

http:/github.com/rishizek/tensorflow-deeplab-v3-plus made for the PASCAL VOC dataset. We would also like to thank Mr. Anurag Bihani for his assistance in creating MudrockNet.